\documentclass[conference]{IEEEtran}
\IEEEoverridecommandlockouts
\usepackage{cite}
\usepackage{soul}
\usepackage{amsmath,amssymb,amsfonts}
\usepackage{algorithmic}
\usepackage{graphicx}
\usepackage{textcomp}
\usepackage{xcolor}
\usepackage{comment}
\usepackage{booktabs} 
\usepackage{algorithm} 
\usepackage{pifont}
\newcommand{\xmark}{\ding{55}}%
\usepackage[autostyle, english = american]{csquotes}
\MakeOuterQuote{"}
\IEEEoverridecommandlockouts

\def\BibTeX{{\rm B\kern-.05em{\sc i\kern-.025em b}\kern-.08em
    T\kern-.1667em\lower.7ex\hbox{E}\kern-.125emX}}

\makeatletter
\newcommand{\linebreakand}{%
  \end{@IEEEauthorhalign}
  \hfill\mbox{}\par
  \mbox{}\hfill\begin{@IEEEauthorhalign}
}
\makeatother

\begin{document}

\title{Exploring Causality for HRI: A Case Study on Robotic Mental Well-being Coaching}

\author{Micol Spitale$^{1, 2}$, Srikar Babu$^{3}$*, Serhan Çakmak$^{4}$*, Jiaee Cheong$^{1}$ and Hatice Gunes$^{1}$
\thanks{$^{1}$ Department of Computer Science and Technology, University of Cambridge, UK. $^{2}$ Department of Electronic, Information, and Bio-engineering, Politecnico di Milano, Italy. $^{3}$ Department of Electrical Engineering, Indian Institute of Technology Madras. $^{4}$ Department of Computer Engineering, Bogazici University. *Contributed to this work while undertaking a remote visiting studentship with the Cambridge AFAR Lab.
        {\tt\small micol.spitale@polimi.it}}%
}

\maketitle

\begin{abstract}
One of the primary goals of Human-Robot Interaction (HRI) research is to develop robots that can interpret human behavior and adapt their responses accordingly. Adaptive learning models, such as continual and reinforcement learning, play a crucial role in improving robots' ability to interact effectively in real-world settings. However, these models face significant challenges due to the limited availability of real-world data, particularly in sensitive domains like healthcare and well-being. This data scarcity can hinder a robot’s ability to adapt to new situations.
To address these challenges, causality provides a structured framework for understanding and modeling the underlying relationships between actions, events, and outcomes. By moving beyond mere pattern recognition, causality enables robots to make more explainable and generalizable decisions.
This paper presents an exploratory causality-based analysis through a case study of an adaptive robotic coach delivering positive psychology exercises over four weeks in a workplace setting. The robotic coach autonomously adapts to multimodal human behaviors, such as facial valence and speech duration. By conducting both macro- and micro-level causal analyses, this study aims to gain deeper insights into how adaptability can enhance well-being during interactions. Ultimately, this research seeks to advance our understanding of how causality can help overcome challenges in HRI, particularly in real-world applications.

\end{abstract}


\section{Introduction}

Research on human-robot interaction (HRI) mainly focuses on the study of
interactions between humans and robots \cite{axelsson2022participant,spitale_2022_review}. 
HRI research typically involves utilising different methodological approaches in order to evaluate and understand the impact of interactive robotic systems and interaction techniques \cite{introduction_lee_2023}.
A crucial factor in the field of HRI is that the successful deployment of robots that collaborate with humans in real-world environments must be addressed by systems that understand not only the environment objects and features, but are also able to learn from human affective states and behavioural interactions \cite{spitale_2022_review}.
%
To achieve this, robotic systems have been increasingly equipped with learning-based models to perceive the environment, generate appropriate behaviour, and adapt their responses according to human needs. 
%

One of the main challenges lies in the limited availability of real-world data that reflects the complexity of interactions. 
In addition to this, gathering such data is highly resource-intensive, requiring extensive human-robot interaction trials across various scenarios, user groups, and environmental conditions.

%
%
The limited availability of real-world interaction data can hinder adaptive models' ability to learn and generalize their behaviors to new situations, potentially leading to underperformance in robotic systems. For example, a reinforcement learning model trained on limited interaction data might struggle to recognise a user's micro-facial expressions, such as a slight frown indicating frustration, or fail to adapt when a user suddenly changes their tone of voice, leading to misunderstandings or an ineffective response from the robot. 
%
%

One promising avenue to address these limitations is the integration of causal reasoning into adaptive models. Causality \cite{gershman2017reinforcement} offers a structured approach to understanding and modeling the underlying relationships between actions, events, and outcomes, enabling robots to move beyond mere pattern recognition toward more explainable and generalizable decision-making. 
Early attempts at leveraging causality to improve robotic capabilities have shown promise~\cite{castri2024ros,diehl2022did}.
However, there is still a lack of research investigating how causal-based tools can be used to improve HRI research, particularly in understanding the causal interplay between human behaviours and robotic actions in real-world interactions.

This work seeks to contribute to bridging this gap by conducting a \textbf{novel exploratory causality-based analysis} applied to a case study in which a robotic coach autonomously adapts to the coachee’s multi-modal behaviours (facial valence and speech duration) while delivering positive psychology exercises over four weeks in a workplace setting \cite{spitale2023vita}. 


\section{Related Work} 


\subsection{Causality in HRI}
%

\begin{table}[htb!]
    \centering
    \small
       \label{tab:causal_works_HRI}
    \caption{An overview of causality works in HRI. }
    \begin{tabular}{ll}
    \hline
      \textbf{Study}   & \textbf{Task} \\
      \hline
       Lee \textit{et al.} \cite{lee_causal21,lee2023scale} 
       & Robot manipulation policies\\
      
      Swamy \textit{et al.} \cite{swamy2022causal} 
      &Simulated control tasks \\
      
      Diehl \textit{et al.} \cite{diehl2022did}
      & Explaining robot failures\\
      
      Frederiksen \textit{et al.} \cite{frederiksen2020causality}
      & Reaction coordination and \\
      &perceived affective impact \\
      
      Castri \textit{et al.} \cite{castri2024experimental}
      & Onbaord spatial interactions \\
      
      Han \textit{et al.} \cite{han2022causal}
      &Robot intent communication\\
      
      Edstrom \textit{et al.} \cite{edstrom2023robot}
      &Robot understanding\\
       \hline
    \end{tabular}

\end{table}

%
%
Causal inference methods have been shown to improve the robustness of robot policies learned from human demonstrations~\cite{swamy2022causal, swamy2022sequence}, explain robot failures~\cite{diehl2022did}, learn robot intent communication~\cite{han2022causal, gao2020joint} and even discover the world dynamics from human interactions~\cite{castri2023continual, edstrom2023robot}.
Frederiksen \textit{et al.} focused on whether there is a causality between reaction coordination and
perceived affective impact on a robot
\cite{frederiksen2020causality}.
Lee \textit{et al.} have further provided an overview of causal inference methods that can be used for human-robot interaction (HRI) research \cite{introduction_lee_2023}.
However,  no existing research has investigated how causality-based tools or causal discovery methods can identify the relationships between various human behavioral cues (e.g., facial expressions or speech features) and well-being or  robot perception measures in the context of human-robot interaction.


\subsection{Robotic mental well-being coaching}
Robotic well-being coaching aims to support people maintain their mental well-being by autonomously delivering psychologically-proven practices. Coaching practices include positive psychology exercises \cite{spitale2023robotic,jeong2023deploying}
meditation activities \cite{axelsson2023robotic}. 
and cognitive-behavioural therapy \cite{scoglio2019use}. 
For example, Jeong et al. \cite{jeong2023deploying} investigated the impact of a robotic coach's role on well-being therapy by comparing its behaviour in assistant, coach, and companion roles. 
%
Adaptive robotic systems \cite{axelsson2023adaptive, spitale2023vita} have further enhanced this field by adapting interactions based on user-specific needs, using multimodal inputs such as speech and facial expressions to tailor robotic interactions.
However, challenges remain, including ensuring the reliability of adaptive behaviours in diverse contexts, and addressing ethical concerns around privacy and data usage \cite{axelsson2023robotic}. 
These initiatives highlight the significance of developing human-centered robotic coaches that build trust, foster collaboration, and enhance user engagement to optimize their support for mental well-being.

\section{Robotic Well-being Coaching Interactions Dataset}


In our previous work \cite{spitale2023vita}, 
we developed an autonomous and adaptive robotic system coach, named VITA, that was evaluated in a 4-week study in which the robotic coach delivered positive psychology exercises \cite{spitale2023vita}. 
%
Details are provided in \cite{spitale2023vita}, here we only provide a summary relevant for the focus of this paper.

\begin{table*}[htb!]
    \centering
    \small
    \label{tab:quest}
        \caption{Questionnaires administrated to participants during the user study used in the macro-analysis.}
    \begin{tabular}{lll}
    \hline
      \textbf{Acronym}   & \textbf{Name} & \textbf{Measure} \\
      \hline
       PANAS  & Positive and Negative Affect Schedule \cite{crawford2004positive}&  \textit{positive} and \textit{negative} affect to evaluate their emotional state\\
       ROSAS &Robotic Social Attributes Scale \cite{carpinella2017robotic}&  the robot’s \textit{warmth}, \textit{competence}, and \textit{discomfort}\\
       WAI-SR & Working Alliance Inventory Short Revised \cite{munder2010working} &  \textit{task}, \textit{goal}, and \textit{bond} to assess the therapeutic alliance with the robot\\
       \hline \\
    \end{tabular}

\end{table*}


\begin{table}[htb!]
    \centering
    
    \caption{Features extracted from audio-video recording of the study.}
    \label{tab:features}
    \begin{tabular}{lllll}
    \hline
      \textbf{Feature}   & \textbf{Description} & \textbf{Type} & \textbf{Macro} & \textbf{Micro} \\
      \hline
       AU1&inner brow raiser& visual & \checkmark& \checkmark\\
       AU2&outer brow raiser& visual& \checkmark& \checkmark\\
       AU4&brow lowerer& visual& \checkmark& \checkmark\\
       AU5&upper lid raiser& visual& \checkmark& \checkmark\\
       AU6&cheek raiser& visual& \checkmark& \checkmark\\
       AU7&lid tightener& visual& \checkmark& \checkmark\\
       AU9&nose wrinkler& visual& \checkmark& \checkmark\\
        AU10 & upper lip raiser & visual& \checkmark& \checkmark\\
        AU12&lip corner puller)& visual& \checkmark& \checkmark\\
        AU14 &dimpler& visual& \checkmark& \checkmark\\
        AU15&lip corner depressor&visual& \checkmark& \checkmark\\
        AU17 &chin raiser& visual& \checkmark& \checkmark\\
        AU20&lip stretcher& visual& \checkmark& \checkmark\\
        AU23&lip tightener& visual& \checkmark & \checkmark\\
        AU25 &lips part& visual& \checkmark & \checkmark\\
        AU26 &jaw drop& visual& \checkmark & \checkmark\\
        AU28 &lip suck& visual& \checkmark & \checkmark\\
        AU45 &blink& visual& \checkmark & \checkmark\\
       \hline
        Pitch && audio& \xmark & \checkmark\\
        Loudness&& audio& \checkmark& \checkmark\\
        Alpha Ratio&& audio& \checkmark& \checkmark\\
        Hammerberg index&& audio& \checkmark& \checkmark\\
        Spectral Flux&& audio& \checkmark& \checkmark\\
         Speech Duration && audio& \xmark & \checkmark\\
         Silence Duration && audio& \xmark & \checkmark\\
        \hline \\
    \end{tabular}
    
\end{table}

We evaluated the VITA system by conducting a longitudinal study that was conducted at a tech consulting company located in Cambridge involving 17 coachees. The study's design, experimental protocol, and consent forms received approval from the Ethics Committee of the Department of Computer Science and Technology, University of Cambridge. The 17 participants interacted with the VITA-based robotic coach that delivered four positive psychology exercises over four weeks as detailed in \cite{spitale2023vita}. 
The robotic coach was equipped with various perception modules including facial expression recognition model, speech features extractor, and interaction rupture detection \cite{spitale2023longitudinal} (i.e., recognising whether the user displays cues of discomfort or the robot makes errors like interrupting the user when speaking). We recorded the sessions with coachees and we gathered both quantitative and qualitative data including interaction logs (i.e., transcriptions of the coachees' speech, the robot’s responses, and outputs from the reinforcement learning models) and questionnaires, namely PANAS, ROSAS, WAI-SR, as reported in Table \ref{tab:quest}.

\begin{figure*}[htb!]
    \centering
    \includegraphics[width=\linewidth]{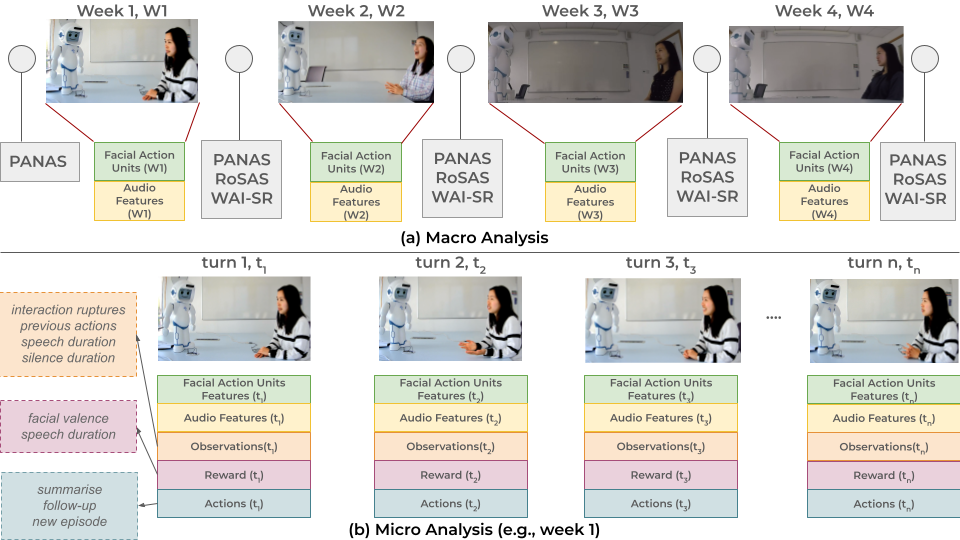}
    \caption{Macro (a, top) and micro (b, bottom) analysis representations.}
    \label{fig:study}
\end{figure*}
%
%
%
%
%
%
The QTrobot by LuxAI S.p.A was used as the robotic platform for delivering positive psychology exercises due to its flexibility (ROS-based) and suitability for this specific application, as demonstrated in previous research \cite{spitale2023robotic, axelsson2023robotic, spitale2022socially}. The QTrobot features a screen face, an RGB-D camera, and a microphone array, enabling real-time computational analysis \cite{spitale2022socially}. 

At each interaction turn $t$—where the robotic coach poses questions, and the participant responds 
—the state of the robotic coach's environment is represented by the observation variable $s_t$. The $s_t$ was a 11-element vector which included interaction rupture prediction (present or absent), the current well-being exercise (savoring, gratitude, accomplishment, or "one door closes, one door opens"), speech features (speech and silence duration), and previous actions.
%
The robotic coach can choose an action $a_t$ at each turn, allowing it to advance the conversation by asking follow-up questions (e.g., "How does this event make you feel?"), summarising the participant’s responses, or initiating a new discussion (e.g., "Can you tell me another thing you were grateful for during the past week?"). 

The primary objective was to learn a conversational policy $\pi$: $s_t$ $\rightarrow$ $a_t$, enabling the robotic coach to effectively deliver the well-being practice. 
Thus, the reward was defined based on the coachees' behavioural cues, specifically combining speech duration and facial valence (i.e., positive or negative emotional value expressed in a person's facial expressions). 
We included facial valence as a reward component since it provides insight into the coachee's emotional state. 
Speech duration was used because, as highlighted in \cite{spitale2023vita}, the human coach adjusted the dialogue flow according to the amount of information shared by the coachee (see Figure \ref{fig:study}).
We extracted behavioural features from the audio-visual data gathered, 
focusing on facial and audio features because previous research has identified them as key indicators of robotic coaching interactions \cite{spitale2023vita, spitale2023longitudinal}.
To process the video recordings, we used the OpenFace 2.2.0 toolkit \cite{openface2018}, extracting the presence and intensity of facial action units (FAUs) to capture coachees' facial cues (as reported in Table \ref{tab:features}), resulting in a total of 17 facial features. 
While, we analysed the audio recordings using the openSMILE toolbox \cite{eyben2010opensmile} to extract interpretable speech features, specifically loudness and pitch, as depicted in Table \ref{tab:features}, resulting in a total of 7 features. 
From these, we derived additional high-level features, such as the duration of the coachees' speech and silence, by processing the audio recordings with the HuggingFace library\footnote{https://huggingface.co/pyannote/speaker-diarization}. Speech diarization was applied to distinguish between the coachee's and the robotic coach's speech, allowing us to calculate the duration of each speaker's audio segments separately.



\section{Methodology}
\label{SCM}
This section outlines the research questions of this study and the causality analysis, which includes a macro-level approach using structural equation modeling and a micro-level analysis conducted through fast causal inference.

\subsection{Research questions}
%
\textbf{RQ1}:How do non-verbal cues influence coachees' perception of a robotic coach, and how does this perception impact the effectiveness of the coaching in promoting their mental well-being?
To investigate this, we conducted a \textbf{macro analysis} that focused on session-level data, which includes aggregated information collected during each coaching session. 
This data encompasses facial and auditory non-verbal cues displayed by the coachees, and overall working-alliance (WAI), perception toward the robots (RoSAS), and mental well-being levels (PANAS) throughout the sessions. This broader perspective allowed us to identify patterns and trends that could inform adjustments to the robot's behaviour.

\textbf{RQ2}: Which are the most influential non-verbal cues that might improve the adaptive capabilities of robotic coaches?
To address this RQ, we conducted a \textbf{micro analysis} that focused on conversational turn-level data. This type of data includes detailed information for each coach-coachee turn in the conversation, such as the specific non-verbal cues displayed by the coachee (e.g., facial expressions, gestures, and body language) and the corresponding actions taken (e.g., summarise what the coachee said) from the robotic coach. 

This dual approach of conducting macro and micro analyses (as shown in Figure \ref{fig:study}) provides a comprehensive framework for developing an adaptive robotic coach. The macro-level analysis enables us to capture overarching patterns across entire sessions, such as how non-verbal cues correlate with overall well-being and alliance-building over time. Meanwhile, the micro-level analysis offers a fine-grained understanding of how individual conversational turns influence immediate interaction dynamics like engagement in the coaching practice. 
Together, these insights would allow for a more nuanced adjustment of the robot's behaviour, balancing long-term interaction goals with real-time conversational feedback, ultimately enhancing the coach's adaptability and effectiveness.

\subsection{Macro-level Analysis: Structural Equation Modeling} \label{sec:macro_model}

We employed Structural Equation Modeling (SEM) \cite{ullman2012structural} to analyse causal relationships between various performance metrics and features captured from robot-coachee interactions during mental well-being coaching sessions. The primary objective is to understand how different facial and audio features influence mental well-being outcomes, PANAS and WAI, coachee's perception towards the robotic coach, i.e., ROSAS, measured after each coaching session.
%
%
The model defines how observed variables $\mathbf{X}$ relate to latent constructs ($\xi$ for exogenous and $\eta$ for endogenous latent variables). 
%
The linear relationships are modeled as: 
\begin{equation}
    \mathbf{X} = \Lambda_{x} \xi + \epsilon_{x}; 
    \mathbf{Y} = \Lambda_{y} \eta + \epsilon_{y}
\end{equation}

where $\mathbf{X}$ and $\mathbf{Y}$ are vectors of observed variables, $\xi$ and $\eta$ represent latent vairbales (exogenous and endogenous), $\Lambda_{x}$ and $\Lambda_{y}$ are factors loadings (coefficients indicating the relationship between observed variables and latent variables), $\epsilon_x$ and $\epsilon_y$ are measurement errors associated with observed variables. Thus, we can model how latent variables like pitch (extracted from audio features) affect ROSAS scores through $\epsilon$ .

 \begin{figure}
     \centering
     \includegraphics[width=\linewidth]{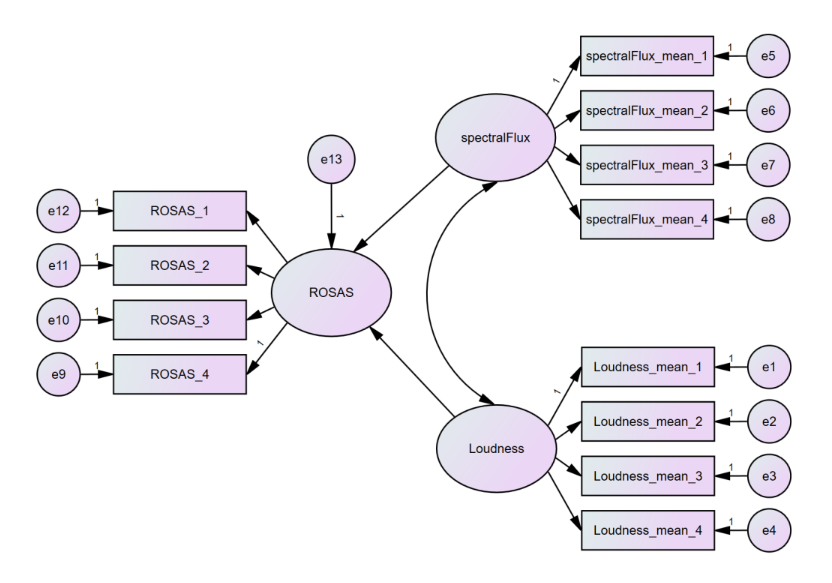}
     \caption{A sample SEM model showing the relation between ROSAS metric and features - spectal flux and Loudness.}
     \label{fig:sample_SEM}
 \end{figure}

%
%
SEM typically employs Maximum Likelihood Estimation (MLE) \cite{ullman2012structural}, where the goal is to minimize the difference between the observed covariance matrix $S$ and the model-implied covariance matrix $\Sigma(\theta)$, where $\theta$ represents the parameters (e.g., factor loadings, path coefficients):
\begin{equation}
    L(\theta) = (S - \Sigma(\theta))^T W (S - \Sigma(\theta)).
\end{equation}

The likelihood function $L(\theta)$ is minimized to estimate $\theta$, subject to the assumption that the data follow a multivariate normal distribution.

\subsection{Micro-level Analysis: Fast Causal Inference (FCI)}
We opted for the FCI algorithm, which is more flexible in handling complex, high-dimensional causal structures and detecting latent variables.
%
%
%
%
%
The Shapiro–Wilk test was performed and showed that the distribution of some features departed significantly from normality (W = 0.35, p-value $< 0.01$). 
%
%
Additionally, we suspected the presence of latent variables that could significantly influence the results. To address these challenges, we employed the Kernel-based Conditional Independence Test (KCIT) 
as our conditional independence test within the FCI algorithm. 
A high-level sketch of FCI \cite{spirtes2000causation}) is
given in Algorithm~\ref{pseudofci}. 

\begin{algorithm}
\caption{The FCI algorithm}
\label{pseudofci}
\fontsize{9pt}{11pt}\selectfont{
\begin{algorithmic}
\REQUIRE Conditional independence information among all variables in
$\mathbf{X}$ given $\mathbf{S}$
\STATE Use Algorithm 4.1 of~\cite{ColomboEtAl12-supp} to
find an initial skeleton ($\mathcal{C}$), separation sets (sepset) and
unshielded triple list ($\mathfrak{M}$);
\STATE Use Algorithm 4.2 of \cite{ColomboEtAl12-supp} 
to orient
v-structures (update $\mathcal{C}$);
\STATE Use Algorithm 4.3 of
to find the final
skeleton (update $\mathcal{C}$ and sepset);
\STATE Use Algorithm 4.2 of
to orient
v-structures (update $\mathcal{C}$);
\STATE Use rules (R1)--(R10) 
to orient as many edge marks as possible (update $\mathcal{C}$);
\RETURN{$\mathcal{C}$, sepset}.
\end{algorithmic}}
\end{algorithm}

%
After identifying the initial skeleton, FCI further processes the graph to uncover potential latent variables that could influence the observed relationships. The refined skeleton is then reoriented using a comprehensive set of additional orientation rules. 
%
%
This approach enables the FCI algorithm to handle real-world data where not all variables are observed, providing a robust framework for causal discovery when data is incomplete or partially observed \cite{spirtes2001anytime}.

\section{Experiments and Results}

\subsection{Macro Analysis Results}
\label{sect:macro_results}

We explored the relationships between various metrics—ROSAS, PANAS, and WAI—and both facial and audio features captured during coaching sessions. 
%
%
%
To evaluate the validity of our models, we used the following performance metrics:
Comparative Fit Index (CFI, i.e., compares the specified model to a null model while accounting for sample size), Root Mean Square Error of Approximation (RMSEA, i.e., assesses model fit per degree of freedom with an emphasis on parsimony), Tucker-Lewis Index (TLI, i.e., evaluates model fit relative to a baseline model while penalizing complexity), and 
Chi-squared/Degrees of Freedom (CMIN/DF, i.e., normalizes the chi-squared value by its degrees of freedom). 
%
%
%
%
The results showed several significant causal relationships:
\subsubsection{Audio Features} For audio features, the ROSAS metric was positively correlated with loudness and the Hammerberg index. This suggests that participants who spoke louder and more clearly during their sessions reported more positive perception toward the robot. In addition, spectral flux—a measure of the variability in speech speed—was positively correlated with ROSAS, implying that faster speech was associated with more positive perceptions of the robotic coach. 
%

\begin{table}
    \centering
    \small
    \caption{Causal relation estimates between metrics and audio features. Estimated Path Coefficient indicates the strength of the relation, positive implies positive correlation and vice-versa (higher in magnitude suggests a stronger relation).}
    \begin{tabular}{|l|l|l|}
    \hline
        \textbf{Metric} & \textbf{Feature} & \textbf{Estimated Path Coefficient}  \\ \hline
        ROSAS & Loudness & 0.681  \\ \hline
        ROSAS & Hammerberg index & 0.436  \\ \hline
        ROSAS & Spectral Flux & 0.439  \\ \hline
        WAI & Loudness & 0.244  \\ \hline
        WAI & Hammerberg index & -0.296  \\ \hline
        WAI & Spectral Flux & 0.139 \\ \hline
    \end{tabular}
\end{table}

\subsubsection{Facial Features} 
We found that certain Action Units (AUs) were strongly related to the well-being metrics. For example, $AU7_{r}$ (Lid Tightening) was negatively correlated with both ROSAS and WAI, indicating that facial expressions associated with tension or stress were linked to more negative perception toward the robot and a weaker therapeutic alliance. Similarly, $AU10_r$ (Upper Lip Raiser) and $AU17_r$ (Chin Raiser) also showed negative correlations with ROSAS and PANAS, suggesting that facial cues of negative emotion or discomfort were associated with more negative perception toward the robotic coach and reduced session well-being.
%
\begin{table}
    \centering
    \small
    \caption{Causal relation estimates between metrics and facial features. Estimated Path Coefficient indicates the strength of the relation.}
    \begin{tabular}{|l|l|l|}
    \hline
        \textbf{Metric} & \textbf{Feature} & \textbf{Estimated Path Coefficient}  \\ \hline
        ROSAS & AU 7 & -1.014  \\ \hline
        ROSAS & AU 10 & -0.137  \\ \hline
        PANAS & AU 17 & -0.063  \\ \hline
        WAI & AU 7 & -0.501  \\ \hline
        WAI & AU 10 & -0.165  \\ \hline
        WAI & AU 15 & -0.164  \\ \hline
        WAI & AU 17 & -0.203 \\ \hline
    \end{tabular}
\end{table}
%
%


\subsection{Micro Analysis Results}
\label{sect:micro_results}

\begin{figure}[htb!]
  \centering
  \includegraphics[width =\columnwidth]{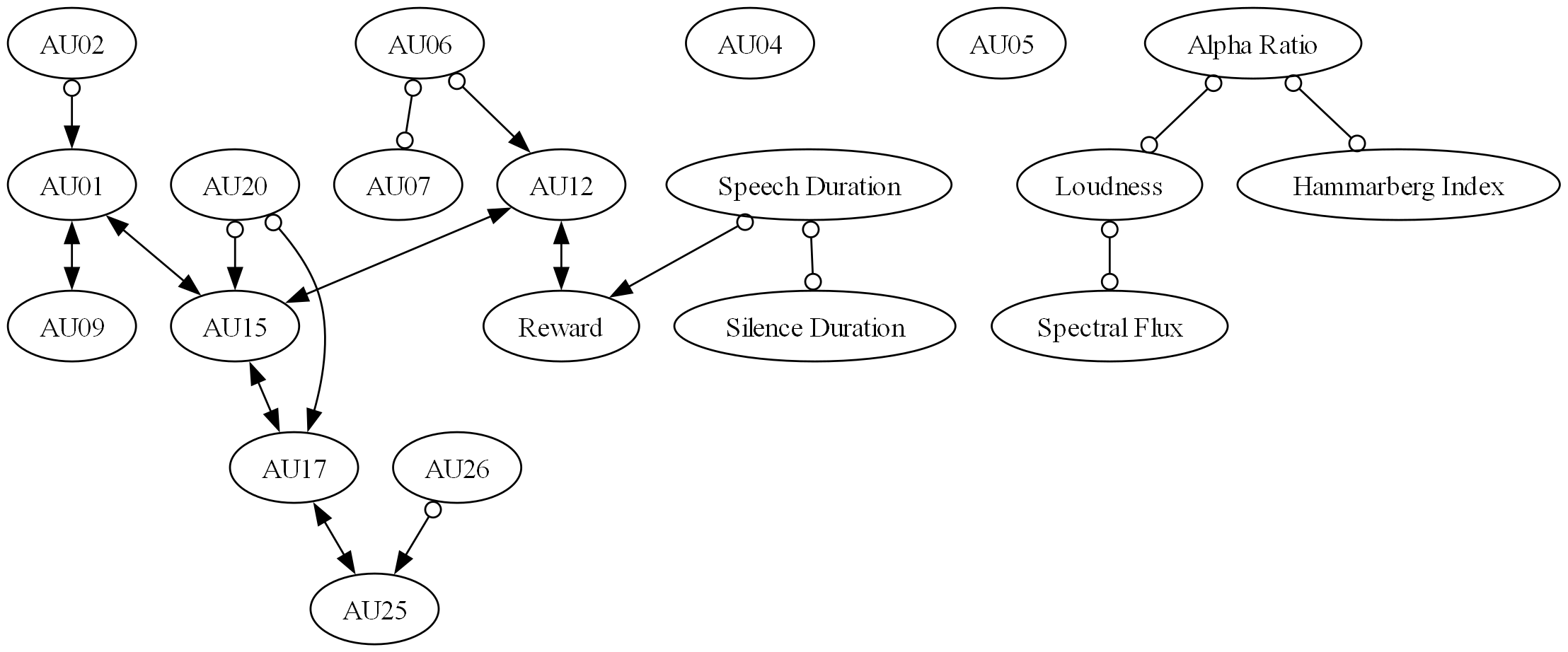}
  \caption{The resulting causal graph obtained using the FCI algorithm. \textit{Legend}:  (1) $X \to Y$, indicating X causes Y; (2) $X \leftrightarrow Y$, suggesting unmeasured confounders affecting both X and Y; (3) X o$\rightarrow$ Y, meaning either X causes Y or there are unmeasured confounders affecting both; and (4) X o-o Y,  representing several possibilities, such as X causing Y, Y causing X, or the influence of latent confounders from both.}
  \label{fig:fci}
\end{figure}

In our study, we employed FCI algorithm, coupled with the KCIT, to analyse our dataset. 
To enhance the robustness of our findings and address potential instability in the causal inferences, we combined two complementary strategies: random sampling and a majority vote synthesis as previously done in \cite{ensemble_causal, buhlmann2011statistics, meinshausen2009stabilityselection}.
Specifically, we sampled 80\% of the dataset without replacement, generating 30 different subsets.
Then, each sample underwent analysis using the FCI algorithm, with the significance level threshold set at 0.05. 
The causal graphs generated from these analyses were then synthesised into a final graph through a majority vote approach.


%
As shown in Figure \ref{fig:fci}, the lack of directed edges suggests the presence of latent confounders (e.g., the daily mood of the coachee, cultural and social norms)  complicating the relationships among FAUs, reward, and vocal features. 
%
%
%
%
The graph reveals a structural separation, with AU15 and its associated nodes generally representing negative emotions, whereas the cluster around AU6 and AU12 corresponds to positive emotions. 
%
%
%
Beyond FAUs, `speech duration' also exhibits a potential causal link with reward. Given that speech duration is a component in the reward calculation, this finding aligns with the function’s definition. 
%


\section{Summary \& Limitation}

%
%
Our results indicate that the evaluation of macro and micro analysis via a causal perspective can facilitate and improve our understanding of the causal effect between the \textit{different human behavioural cues} and contextual variables, like human perception.
%
%
%
Across the \textbf{macro-level} causal analysis, 
our results 
demonstrate the complex interplay between vocal and facial features and the perception toward the robotic coach. 
The strong fit of the models, as inferred by the high CFI values and low RMSEA scores, supports the validity of the causal relationships identified. 
%
%
%
%
%
Across the \textbf{micro-level} causal analysis, 
our results demonstrate that there is an underlying causal structure. 
A key limitation is that causal effect estimation or inference is challenging because weonly rely or observe the factual, but not the counterfactual outcomes. 
In this work, we have mainly relied on the Maximum Likelihood Estimation (MLE) method to quantify causal effects. However, MLE is not the only measure that can be used to quantify causal effects. 
Future work can address this gap and extend this methodology to other human-robot interaction setting \cite{castri2024ros,castri2024experimental}.
%
%
%
%

\section*{Acknowledgments}
\noindent
We thank Cambridge Consultants Inc. and its employees for participating in this study. 
\textbf{Funding:} M. Spitale, J. Cheong and H. Gunes have been supported by the EPSRC/UKRI under grant ref. EP/R030782/1 (ARoEQ) and EP/R511675/1. M. Spitale is currently supported by PNRR-PE-AI FAIR project funded by the NextGeneration EU program.  \textbf{Open Access:} For open access purposes, the authors have applied a Creative Commons Attribution (CC BY) licence to any Author Accepted Manuscript version arising.
\textbf{Data access:} Raw data related to this publication cannot be openly released due to anonymity and privacy issues.

\normalsize
\bibliographystyle{IEEEtran}
\bibliography{ref}
\end{document}